\pdfoutput=1

\documentclass[11pt]{article}

\usepackage[final]{acl}

\usepackage{times}
\usepackage{latexsym}

\usepackage[T1]{fontenc}

\usepackage[utf8]{inputenc}

\usepackage{microtype}

\usepackage{inconsolata}

\usepackage{graphicx}

\usepackage{subcaption}
\usepackage{xcolor}
\usepackage{float} 
\usepackage{tabularx} 
\usepackage[most]{tcolorbox}

\title{Nollywood: Let's Go to the Movies!}

\author{John E. Ortega \\
  Northeastern University \\
  \texttt{j.ortega@northeastern.edu} \\
  \And
   Ibrahim Said Ahmad \\
  Northeastern University \\ 
  \texttt{i.ahmad@northeastern.edu} \\
  \AND
  William Chen \\
  Carnegie Mellon University \\
  \texttt{williamchen@cmu.edu}}

\begin{document}
\maketitle
\begin{abstract}
\textit{Nollywood}, based on the idea of Bollywood from India, is a series of outstanding movies that originate from Nigeria. Unfortunately, while the movies are in English, they are hard to understand for many native speakers due to the dialect of English that is spoken. In this article, we accomplish two goals: (1) create a phonetic sub-title model that is able to translate Nigerian English speech to American English and (2) use the most advanced toxicity detectors to discover how toxic the speech is. Our aim is to highlight the text in these videos which is often times ignored for lack of dialectal understanding due the fact that many people in Nigeria speak a native language like Hausa at home.

\end{abstract}

\section{Introduction}
\label{sec:intro}

In the past several decades, there has been a significant amount of research on digital systems pertaining to language. Some state-of-the-art digital language systems, like those based on automatic speech recognition (ASR), are now considered to be on par with humans for high-resource languages like English and Spanish for conversational speech recognition\cite{min-asr-2023}. However, there are still challenges that significantly impact the performance of ASR, such as the recognition of English with accents \cite{Hinsvark2021AccentedSR}. The difficulty in recognizing accented English can be attributed to the diversity in pronunciation, intonation speed, and pronunciation of specific syllables. 

In some countries, low-resource languages can affect how the high-resource language is spoken which poses many challenges from the ASR standpoint of view. For example, one recent study \cite{ngueajio2022hey} shows that more often than not, ASR systems can be non-inclusive. The struggles that one may encounter as an end user are not the focus of this article but provide a backdrop for a problem that we implicitly address: how does the culture and other attributes of an English speaker affect the digital processing of speech and other resources. In this article, we focus on two main sub-fields of digital processing: \textit{speech recognition} (SR) and \textit{toxicity} (TX). 

For our study to be valid and useful for those who speak low-resource languages like Hausa (a low-resource language spoken in Nigeria, Africa), we dive into a Nigerian digital movie genre called: \textbf{Nollywood}. Nollywood is the Nigerian video film industry that emerged within the context of several pre-existing theatre traditions among various ethnic groups of Nigeria \cite{alabi2013introduction}. Nollywood is currently the third largest film industry in the world, and it has generated over 500 million dollars since inception \cite{umukoro2020nollywood}. According to \cite{ezepue2020new}, Nollywood is gaining popularity due to its transformations, which bear resemblances to the processes of gentrification and professionalization. This is formalizing the industry as well as attracting professionals and instigating existing filmmakers to improve on their art. Nollywood has impact not just in Nigeria, but across the African continent and beyond. 


In this paper, we focus on ASR for two countries where English is the official (high-resource) language: Nigeria and the United States of American (USA). These two countries represent a large part of the English-speaking world. Nigeria has a population of more than 200 million\footnote{\url{https://www.census.gov/popclock/world/ni}} while the USA has a population of more than 300 million\footnote{\url{https://en.wikipedia.org/wiki/Demographics_of_the_United_States}}. Both countries generally use English for business and everyday conversation since their official languages are English \cite{danladi2013language}. However, in areas such as New York City in the USA and in most of Nigeria, native-language accents can have an influence on how English is spoken. It is a known fact that ASR systems are known to be faulty when tested with non-native speakers. \cite{benzeghiba2006impact} 

The authors of this work all speak English as it is the high-resource language of their countries. As one of the co-authors speaks Hausa and has first-hand experience, we consider our approach somewhat more inclusive. Moreover, this article brings attention to specific dialectical approaches that could be used in other languages such as Spanish or Chinese (languages which the other two authors speak fluently). We feel that this gives us insights into the problem that other investigators may not have. There is a difference between the accent influence in Nigeria versus that of the USA. In our opinion, English takes the forefront of culture in the USA and has a dominant force, especially in the movie industry. That does not seem to be the same in Nigerian movies such as the Nollywood movies. In this article, we attempt to build a state-of-the-art ASR system for Nigerian English to see if the system performs well when applied to the Nigerian dialect. Additionally, we compare the two languages to determine the amount of toxicity (words that are considered taboo or should not be spoken in formal language).

In order to better describe our experimentation, we divide this article into several sections. First, in Section \ref{sec:related_work} we introduce work that has motivated our experiments and is related to ours. We then cover our approach in more detail for ASR and TX in Section \ref{sec:methodology}. Thirdly, we compare our results in Section \ref{sec:results}. Finally, we provide further analysis and conclude in Section \ref{sec:conclusion} and provide some ideas of future work in Seciton \ref{sec:future}.

\section{Related Work}
\label{sec:related_work}
The article by \citet{Amuda_Boril_Sangwan_Ibiyemi_Hansen_2014} investigates the engineering analysis and recognition of Nigerian English (NE) in comparison to American English (AE) using the UILSpeech corpus. The study utilizes speech audio and video data to analyze speech parameters and their impact on automatic speech recognition (ASR) systems. Data collection includes isolated word recordings and continuous read speech data from Nigerian English speakers, highlighting the linguistic diversity of the region. The research employs techniques such as lexicon extension and acoustic model adaptation to address phonetic and acoustic mismatches between NE and AE, resulting in a 37\% absolute word error rate reduction. The main findings emphasize the importance of tailored lexicons and acoustic models for low resource languages, showcasing the potential for improved ASR performance in dialectal variations. However, the study acknowledges limitations in the generalizability of findings to other low resource dialects and the need for further research on speaker-dependent phonetic patterns.

\citet{Babatunde_2023} presents a novel approach to developing an Automatic Speech Recognition (ASR) system tailored for Nigerian-accented English by leveraging transfer learning techniques on pretrained models, including NeMo QuartzNet15x5 and Wav2vec2.0 XLS-R300M. The research addresses the challenges of recognizing and transcribing Nigerian accents, aiming to ensure equitable access to ASR technologies for individuals with Nigerian accents. The NeMo QuartzNet15x5Base-En model demonstrated promising results with a Word Error Rate (WER) of 8.2\% on the test set, showcasing its effectiveness in handling Nigerian-accented speech data. However, limitations such as the small dataset size and overfitting observed in the Wav2vec2.0 XLS-R300M model were acknowledged. This work contributes to the advancement of ASR systems for African-accented English, emphasizing the importance of inclusivity and accurate transcription in diverse linguistic communities.

\citet{Oluwatomiyin-2022} explores the development of a hybrid translation model for converting pidgin English to the English language, utilizing the JW300 corpus for training and evaluation. The study employs Phrase-based Statistical Machine Translation (PBSMT) and Transformer-based Neural Machine Translation models to enhance translation accuracy. Results indicate that the hybrid model surpasses the baseline NMT model, demonstrating improved performance in translation tasks with the highest BLEU score of 29.43 using the pidgin-pbsmt model. The findings highlight the potential of combining PBSMT and NMT techniques to enhance translation quality for low-resource languages. However, limitations include challenges related to vocabulary size and computational resources, suggesting the need for further research to address scalability issues and optimize the model for broader applications.

\begin{figure*}[ht]
\begin{tcolorbox}
\begin{itemize}
    \item[] \textbf{Sentence 1:} 
    \item[] \textit{Hey, have you heard the latest gist about the party next weekend? It's gonna be lit!}
    \item[] \textbf{Sentence 2:}  
    \item[] \textit{Let's schedule the meeting for October 10th at 2:30 PM.}
    \item[] \textbf{Sentence 3:} 
    \item[] \textit{I'll meet you at the gas station; we can take the freeway to the shopping mall.}
    \item[] \textbf{Sentence 4:} 
    \item[] \textit{The project deadline is tomorrow, and I need to submit my resume to the recruiter.}
\end{itemize}
\end{tcolorbox}
\caption{Four sentences used to create spectrograms for initial comparison between English spoken in Nigeria and the United States of America.}
\label{fig:sentences}
\end{figure*}

An article \citet{Oladipupo_Akinfenwa_2023} examines the phonemic realisation of educated Nollywood artistes in Nigeria and their accent as a normative standard of English pronunciation. It analyzes the pronunciation of various phonemes, comparing them to Received Pronunciation (RP) forms. The study focuses on the competence of educated Nollywood artistes in pronouncing these phonemes, highlighting improvements in the realisation of certain vowels compared to typical Nigerian English accents. The research suggests that these artistes could serve as normative pronunciation models for Nigerian English learners. Additionally, the article discusses the debate surrounding the codification of Nigerian English as a standard for communication and learning in Nigeria, considering the influence of native English models and technology-driven speech practice sources.

\section{Methodology}
\label{sec:methodology}

Our experiments represent two of the most modern tasks currently being investigated in the natural language processing (NLP) field. Recent conferences such as Interspeech\footnote{\url{https://interspeech2023.org/}} and ACL\footnote{\url{https://2023.aclweb.org/}} have including ASR and TX as main focuses in workshops and other publications. While low-resource languages are heavily investigated for tasks such as machine translation (MT), the effect of low-resource language speaker's accents on English is not heavily researched. In this section, we present several state-of-the-art systems for ASR and TX with a focus on dialectal difference between English spoken in movies from Nollywood and Hollywood.

\begin{figure*}[!htp]
    \centering
    \begin{subfigure}[b]{0.45\textwidth}
        \includegraphics[width=\textwidth]{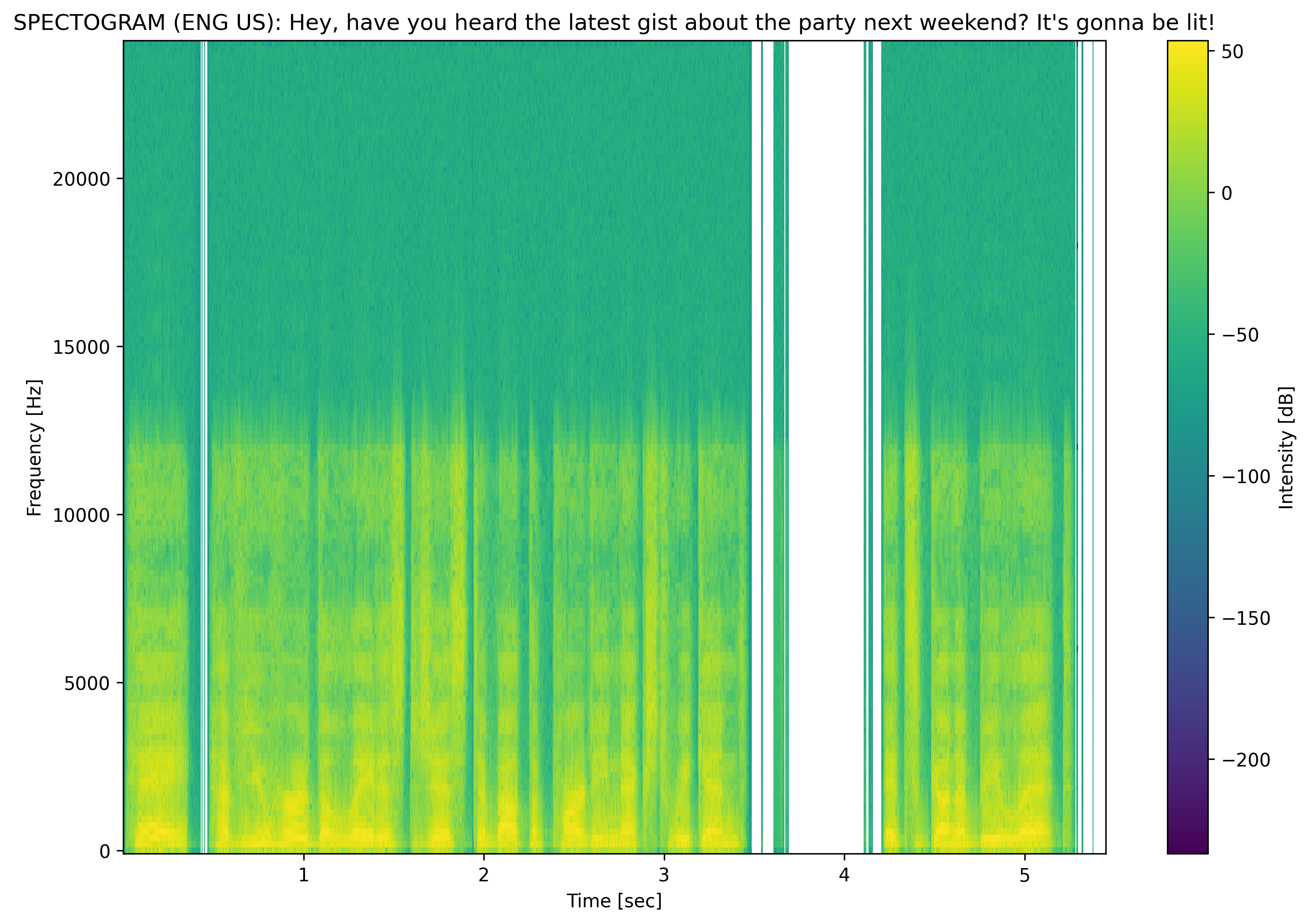}
        \caption{Spectrogram for Sentence 1 (US English) }
        \label{fig:sub1}
    \end{subfigure}
    \hfill
    \begin{subfigure}[b]{0.45\textwidth}
        \includegraphics[width=\textwidth]{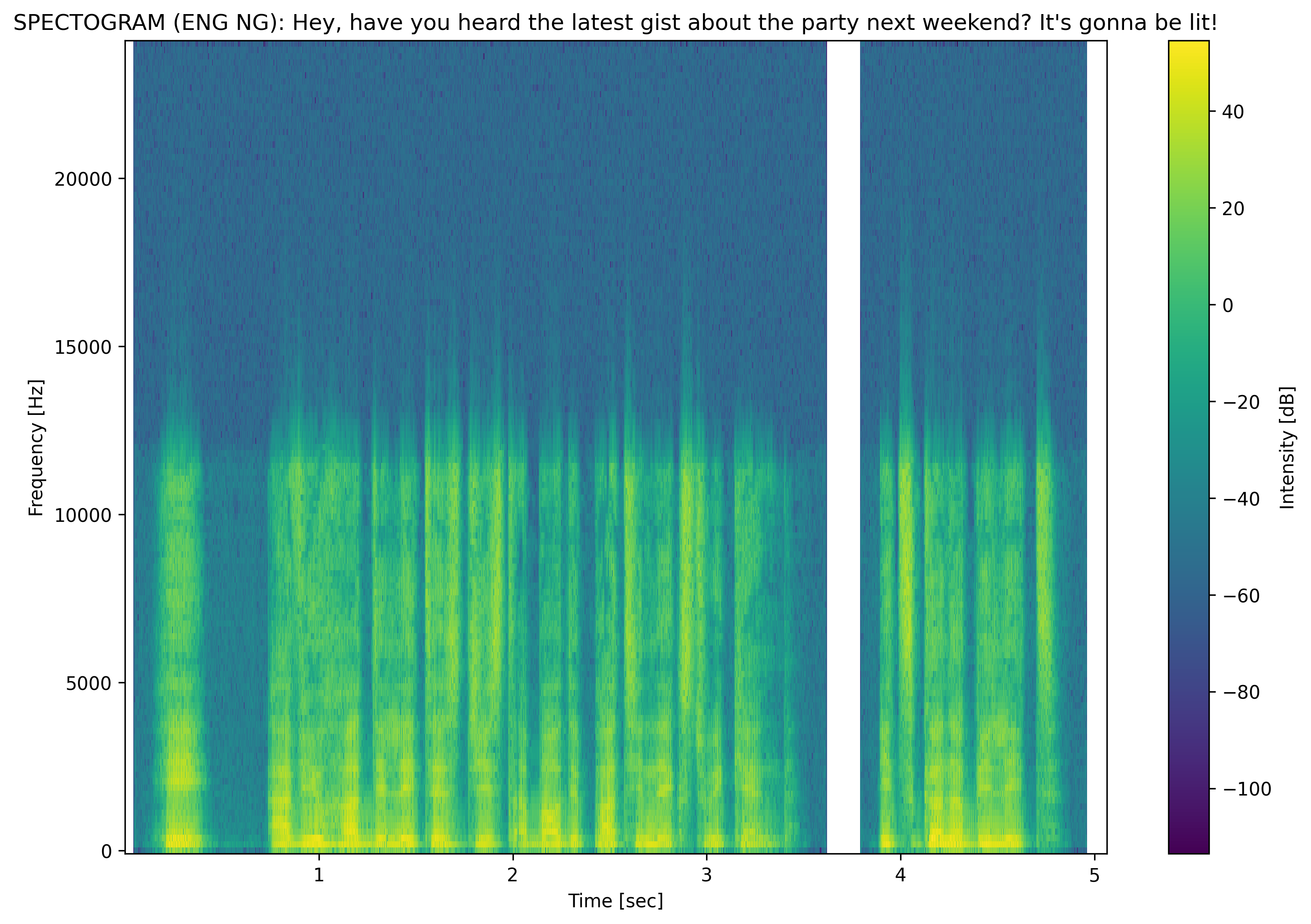}
        \caption{Spectrogram for Sentence 1 (NG English)}
        \label{fig:sub2}
    \end{subfigure}

    \vspace{\baselineskip} 

    \begin{subfigure}[b]{0.45\textwidth}
        \includegraphics[width=\textwidth]{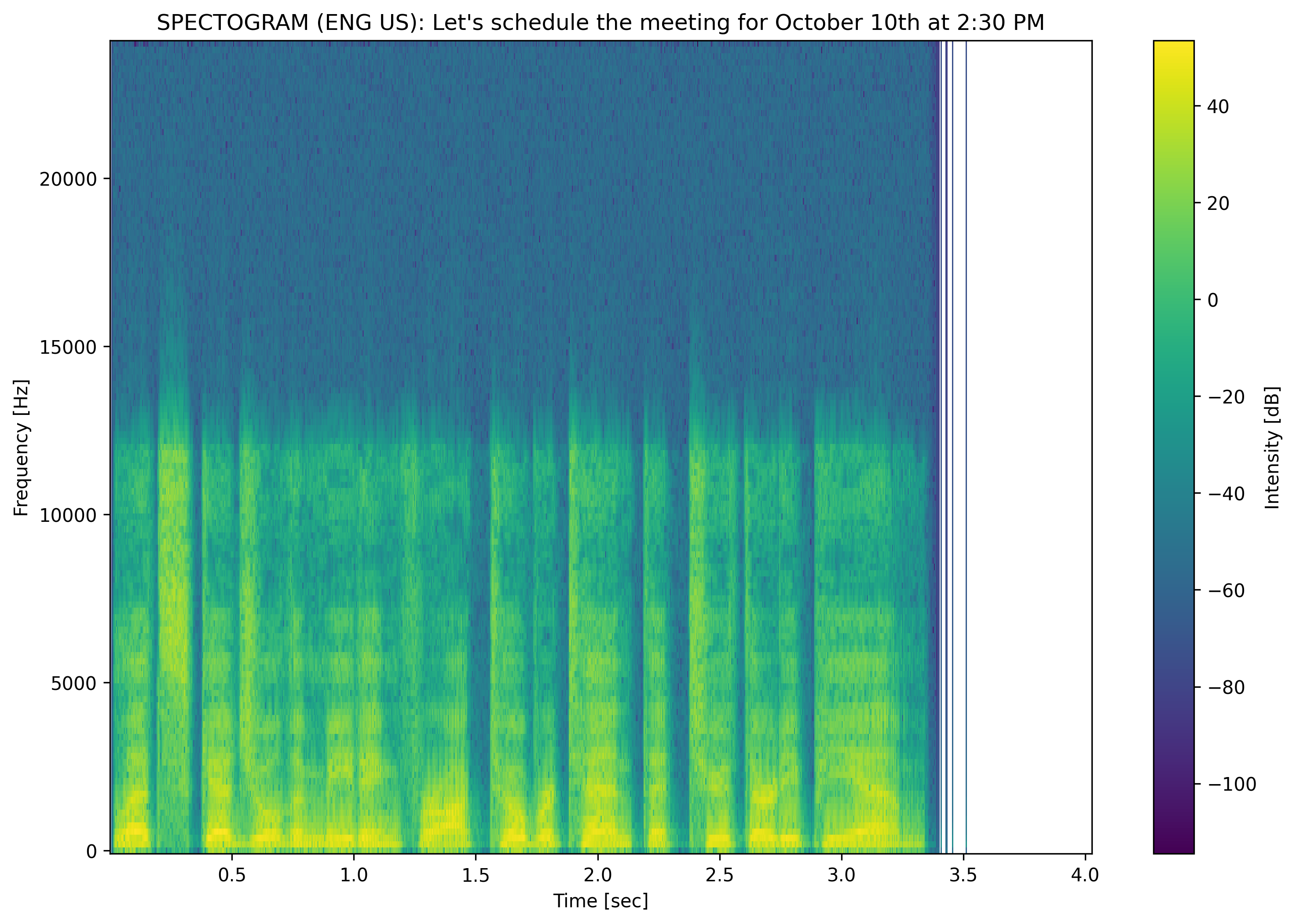}
        \caption{Spectrogram for Sentence 2 (US English)}
        \label{fig:sub3}
    \end{subfigure}
    \hfill
    \begin{subfigure}[b]{0.45\textwidth}
        \includegraphics[width=\textwidth]{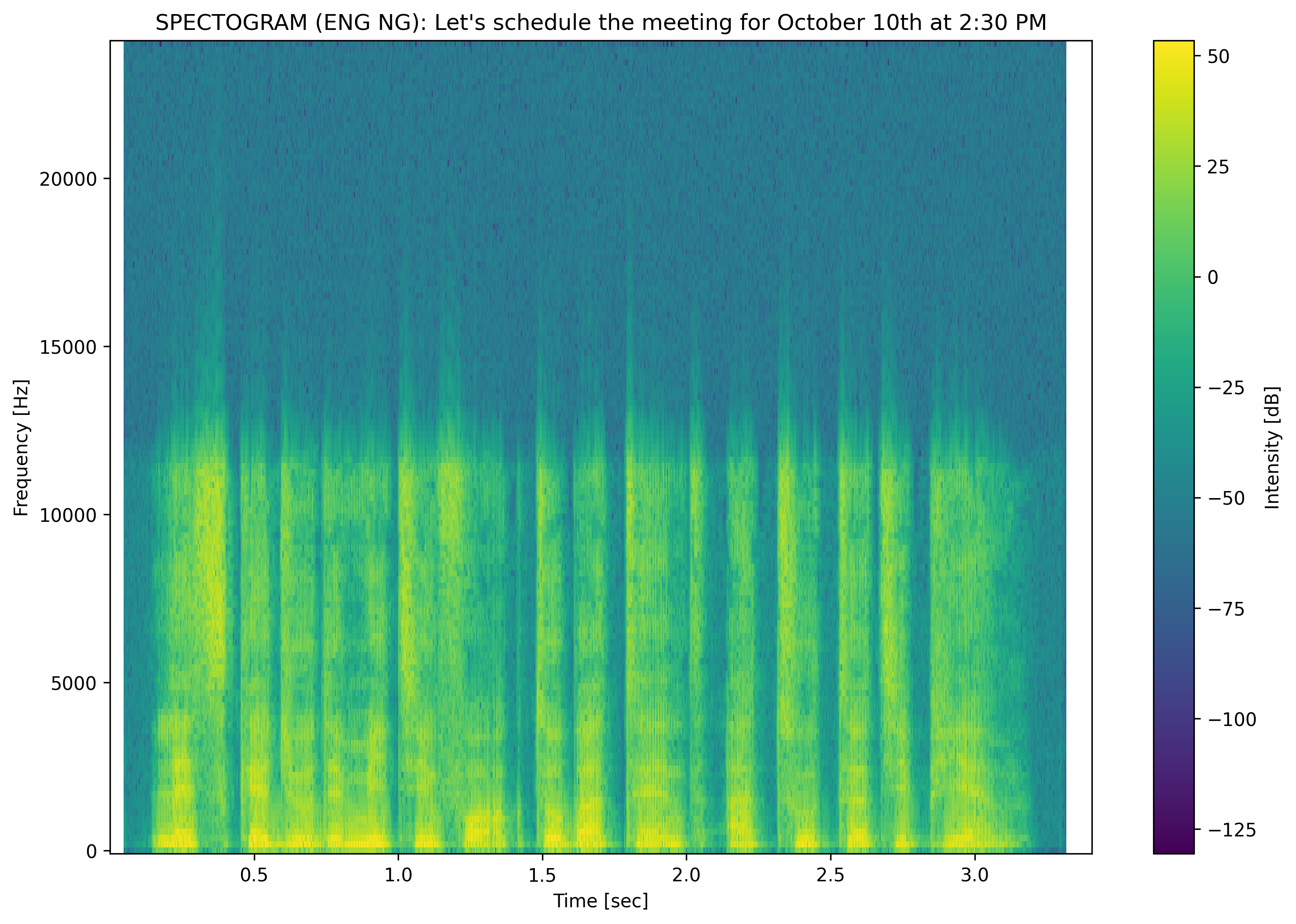}
        \caption{Spectrogram for Sentence 2 (NG English)}
        \label{fig:sub4}
    \end{subfigure}

    \vspace{\baselineskip} 

    \begin{subfigure}[b]{0.45\textwidth}
        \includegraphics[width=\textwidth]{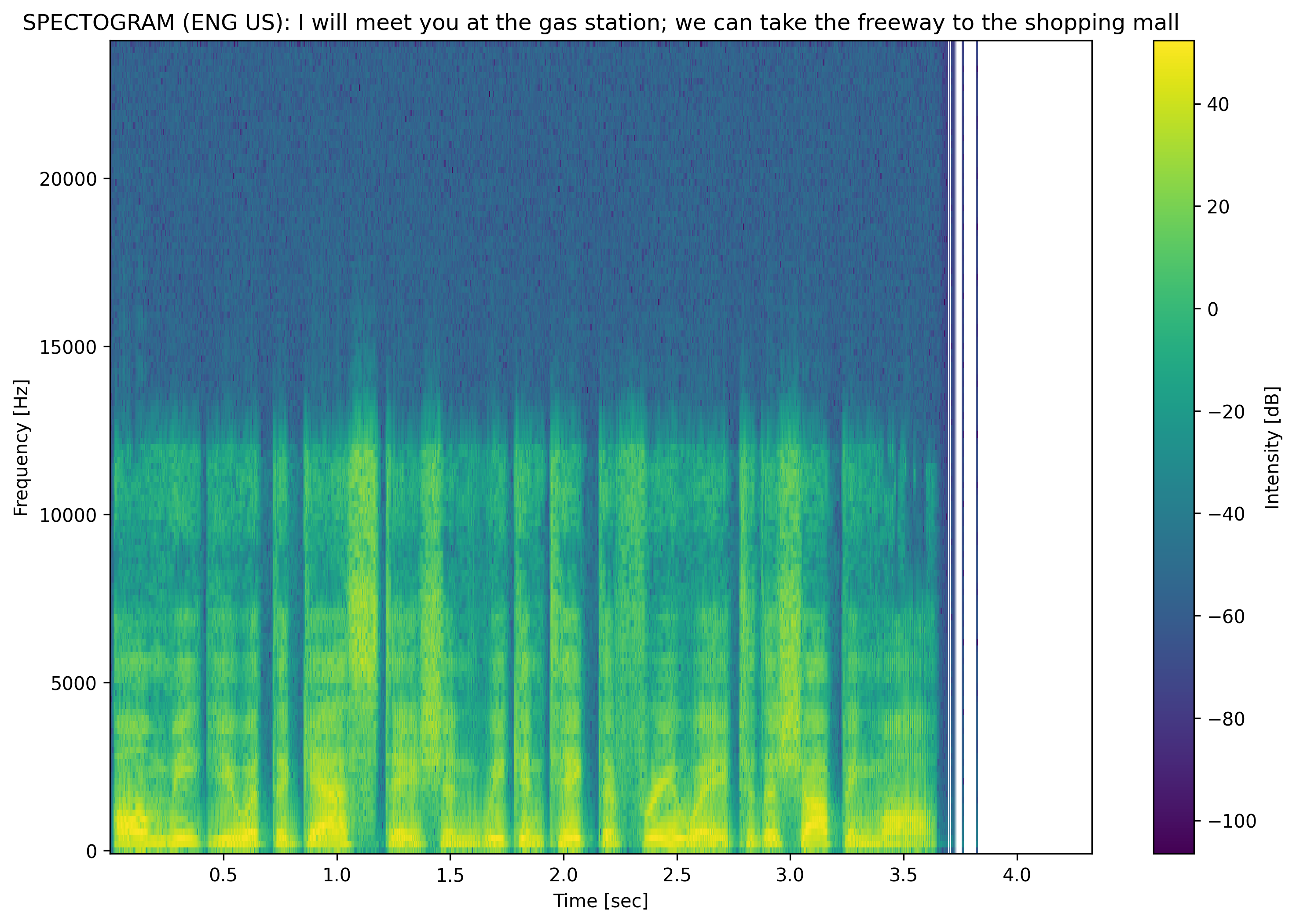}
        \caption{Spectrogram for Sentence 3 (US English)}
        \label{fig:sub5}
    \end{subfigure}
    \hfill
    \begin{subfigure}[b]{0.45\textwidth}
        \includegraphics[width=\textwidth]{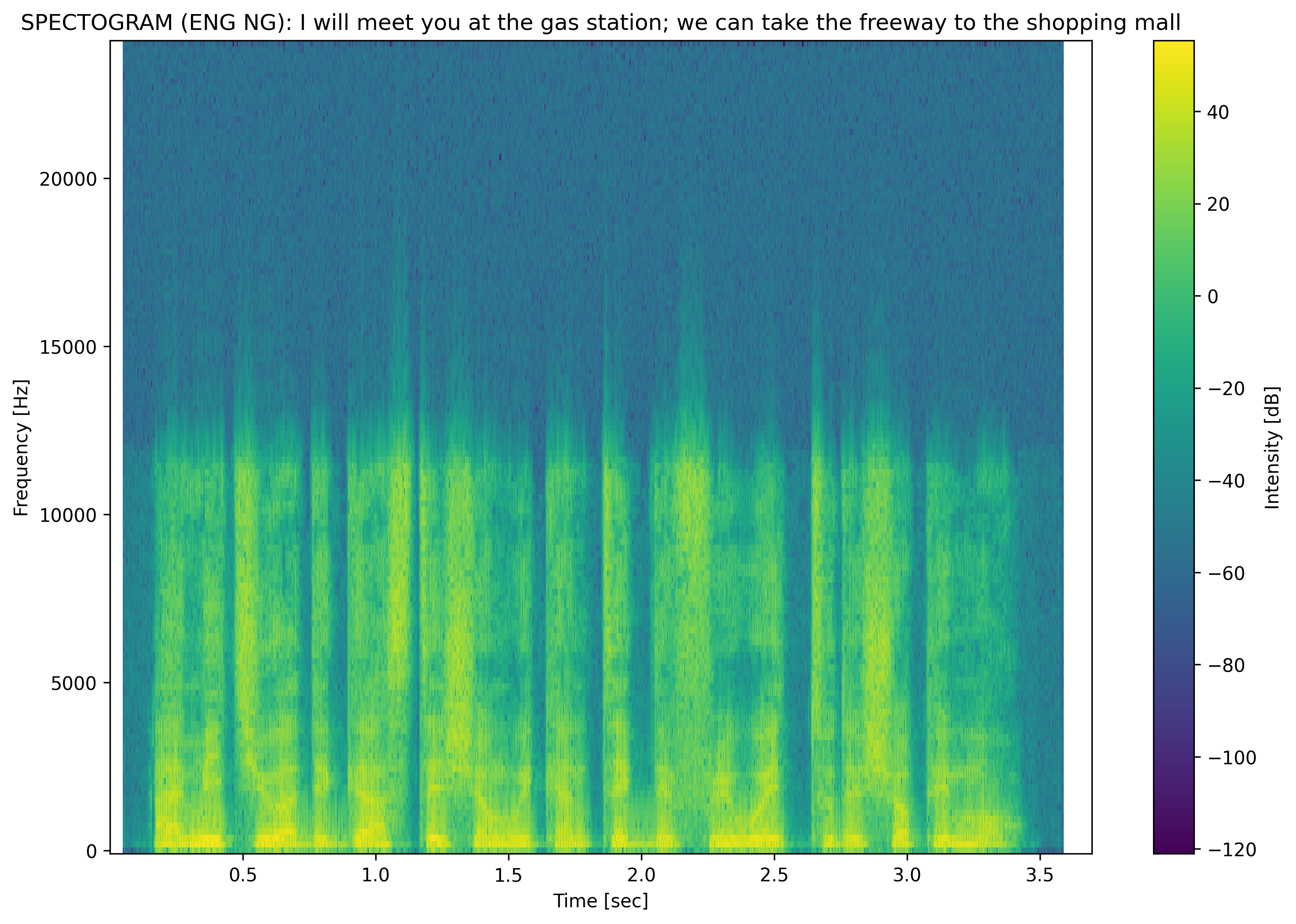}
        \caption{Spectrogram for Sentence 3 (NG English)}
        \label{fig:sub6}
    \end{subfigure}

    \vspace{\baselineskip} 

    \begin{subfigure}[b]{0.45\textwidth}
        \includegraphics[width=\textwidth]{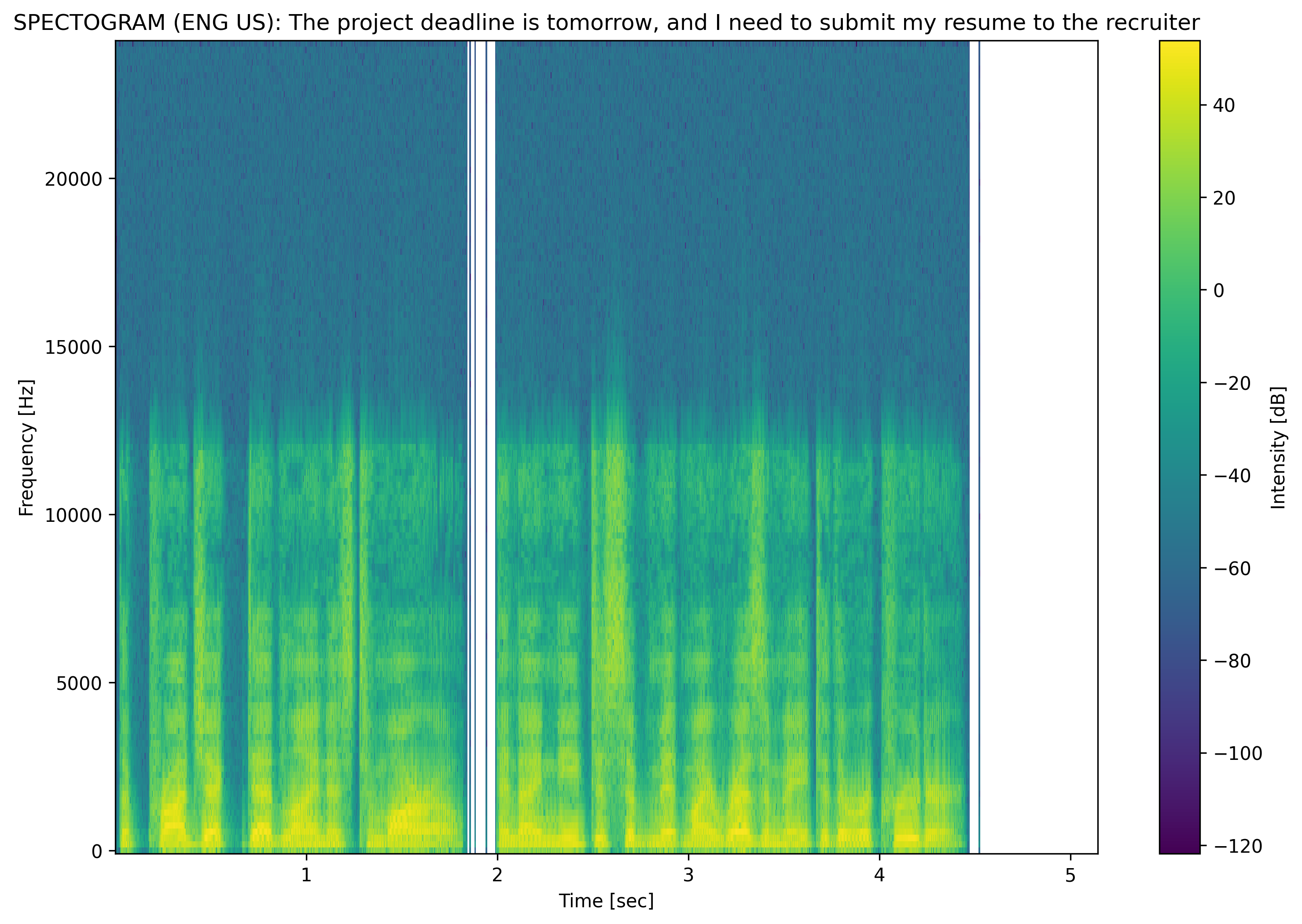}
        \caption{Spectrogram for Sentence 4 (US English)}
        \label{fig:sub7}
    \end{subfigure}
    \hfill
    \begin{subfigure}[b]{0.45\textwidth}
        \includegraphics[width=\textwidth]{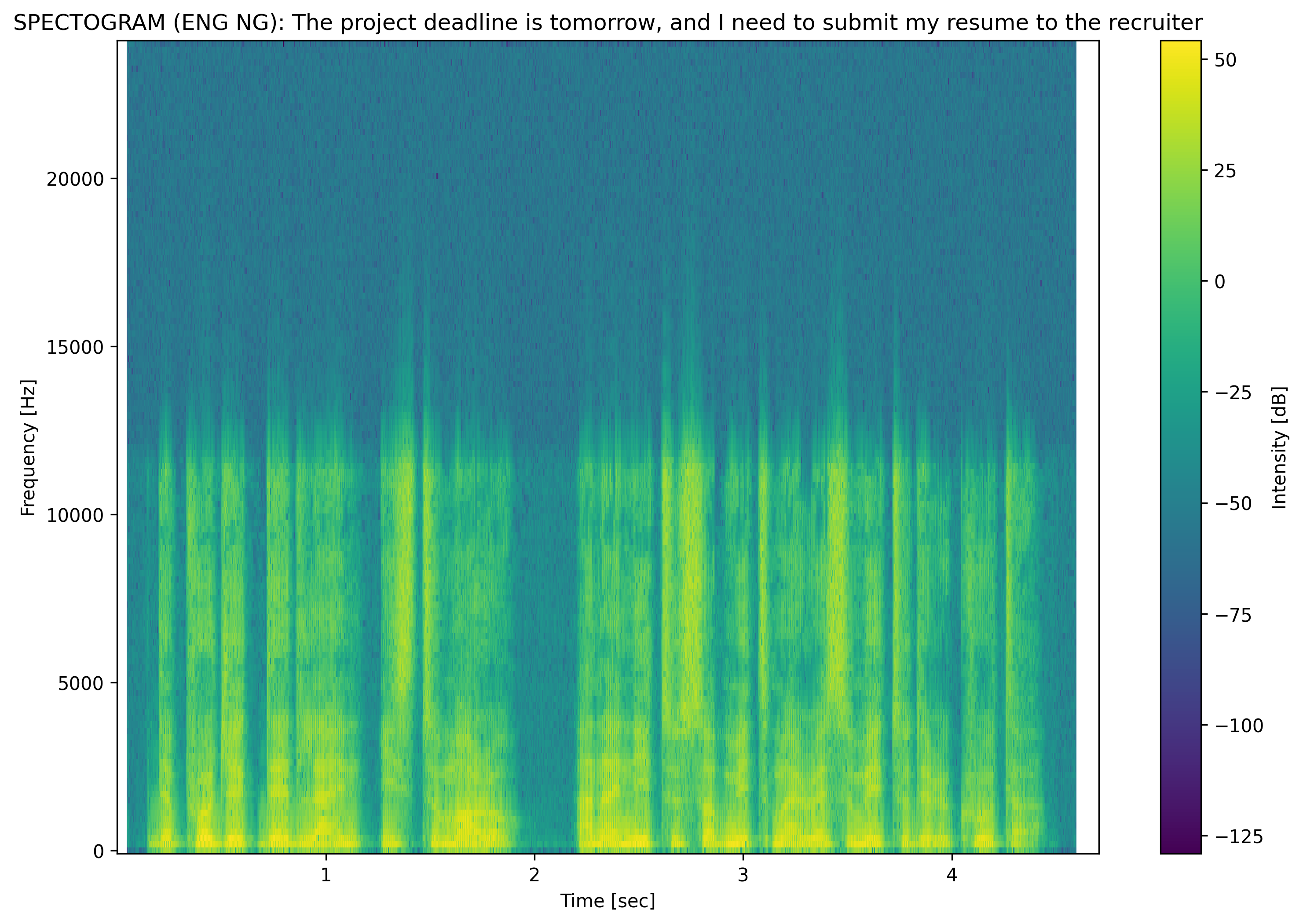}
        \caption{Spectrogram for Sentence 4 (NG English)}
        \label{fig:sub8}
    \end{subfigure}

    \caption{Spectrogram comparison of four sentences in English spoken by speakers from the USA and Nigeria.}
    \label{fig:spectograms}
\end{figure*}

In order to clearly illustrate the need to better identify the difference between the two dialects, we first motivate our experiments visually by comparing identical sentences spoken by two male counterparts: a Nigerian speaker and a USA speaker. We use these examples as ample evidence to show that there is quite a bit of difference between the two for sentences that are not complex. In Figure \ref{fig:spectograms}, we compare and contrast spectrogram samples for the four sentences in Figure \ref{fig:sentences}.

In order to create the spectrograms, we had to find a way to create an audio file (.wav format) that would take as input one of the four sentences from Figure \ref{fig:spectograms}. An online tool called SpeechGen\footnote{\url{https://speechgen.io}} allowed us to perform the task and the output was verified by the authors, Nigerian and USA native speakers.

It is clear that there is a noted difference visually from the generated speech files for Sentences 1 through 4 in Figure \ref{fig:spectograms}. For example, for Sentence 3 the US English (e) and Nigerian English (f) between seconds 2.0 and 2.5 are quite different. The Nigerian speaker seems to have higher frequency and contain more volume. While the audio files where created using a digital ASR system; actual human voice could be more expressive. In this effort, we wanted the system to be equal in order to measure the main digital difference between the two languages as no two humans can be considered to have the same dialect or voice \cite{karpf2006human}. With the notion of difference between the two dialects we present the following steps taken to repeat our experiments.

\subsection{Corpora}
For our experiments to be realistic and capture differences in everyday movies, we assess two major films: (1) the movie Deep Cut\footnote{\url{https://www.youtube.com/watch?v=Xl6ANUHjEtI}} from the Nigerian Nollywood theme and (2) Acrimony\footnote{\url{https://en.wikipedia.org/wiki/Acrimony_(film)}} from the US Hollywood theme. We use these two examples as random picks for movies that could be considered from the nearly the same genres and containing similar topics. We gather the text transcriptions from both movies to assess first \textit{toxicity}. Then, for a more direct evaluation of how well ASR fares on Nigerian corpus, we use the International Corpus of English (ICE)-Nigeria corpus \cite{wunder2010ice}. We did not test on ASR for the USA dialect as it has already been reported on in several conferences.

\subsection{Toxicity}

Toxicity detection is an important challenge in natural language processing (NLP), as highlighted by the latest research in this area \cite{soldaini2024dolma}. For example, \citet{sun2024trustllm} proposed TrustLLM, a framework for trustworthy large language models that incorporates techniques for detecting and mitigating toxic and harmful content.

\subsubsection{Toxicity Metric} 
The Toxicity metric\footnote{\url{https://huggingface.co/spaces/evaluate-measurement/toxicity}} is part of the Evaluate library, a collection of pre-trained models found on HuggingFace for a wide range of evaluation tasks. The library is developed by \citet{vidgen-etal-2021-learning}. The toxicity metric quantifies the harmful or offensive content in text using a pretrained hate speech classification model based on the \textbf{\textit{roberta-hate-speech-dynabench-r4}}. In the model, `hate’ is defined as ``abusive speech targeting specific group characteristics, such as ethnic origin, religion, gender, or sexual orientation.”. It assesses whether a given text is toxic or non-toxic, making it valuable for toxicity detection and bias assessment. It is noteworthy however that context and cultural norms are vital to interpreting toxicity scores.

\subsubsection{Seamless4MT}
In the context of machine translation, frameworks like Seamless4MT from Meta \cite{barrault2023seamlessm4t} employ modern techniques such as large language models and generative approaches to address various challenges, including the potential introduction of toxicity during translation. Seamless4MT is considered state-of-the-art for translation and is deemed stable enough to be used as a reliable tool for detecting toxicity in text.

In this study, we aim to measure the prevalence of toxicity in two films, Deepcut and Acrimony, using Seamless4MT's toxicity detection capabilities. This initial investigation is intended to assist researchers working on dialect detection in Nigeria by providing insights into whether certain words from these movies were found to be more toxic in one dialect compared to another. Both films are rated R for language, indicating an expectation of some toxic language content. The purpose is also to assess the extent to which movies with such a language rating are considered toxic by state-of-the-art toxicity detection techniques like Seamless4MT.




\subsubsection{ETOX}
In our experiments, we mirror Meta's work by using ETOX\footnote{\url{https://github.com/facebookresearch/stopes/tree/main/demo/toxicity-alti-hb/ETOX}}. ETOX is an approach to detect toxic content in text that relies on the \citet{nllbteam2022language} wordlists containing profanities, insults, hate speech terms, and explicit language across 200 languages. These wordlists were created through human translation efforts. The toxicity detection method used by ETOX is based on checking if a given sentence contains any words present in the corresponding language's toxicity wordlist. A word is considered present if it appears as a separate token surrounded by spaces or punctuation.

One advantage of this wordlist-based approach is that it provides transparency, reducing the possibility of encoding biases compared to alternative methods like machine learning classifiers. While classifiers are available for English and some other languages, they cannot be easily applied across hundreds of languages.

ETOX was also utilized by \citet{barrault2023seamlessm4t} to propose a new metric called ASR-ETOX for detecting toxicity in speech data. This metric follows a two-step process: first using an automatic speech recognition (ASR) system to transcribe the audio, and then applying the ETOX toxicity detection module on the transcribed text.

\subsection{Automatic Speech Recognition}
\begin{table*}[!htbp]
    \newcolumntype{C}{>{\centering\arraybackslash}X}
    \centering
    \caption{Toxicity results}
    \label{tab:toxicity-table}
    \begin{tabularx}{\textwidth}{CCCCC}
        \hline
            & \textbf{Deepcut} & \textbf{Acrimony} & \textbf{ICE Spoken} & \textbf{ICE Written} \\
        \hline
        ETOX & 2.08\% & 3.35\%  & <1\%  & <1\% \\
        Evaluate & 1.30\% & 2.16\%  & <1\%  & <1\% \\
        \hline
    \end{tabularx}
\end{table*}

ASR tools are plentiful for US English; however, here our main goal is to test the validity of the latest technique for a dialect in English that is not in mainstream research: Nigerian. In order to do that, we use the largest corpus we could find for training an ASR model on English text with a Nigerian accent: \textbf{ICE}.

\begin{figure}[!htbp]
    \centering
    \includegraphics[width=0.48\textwidth]{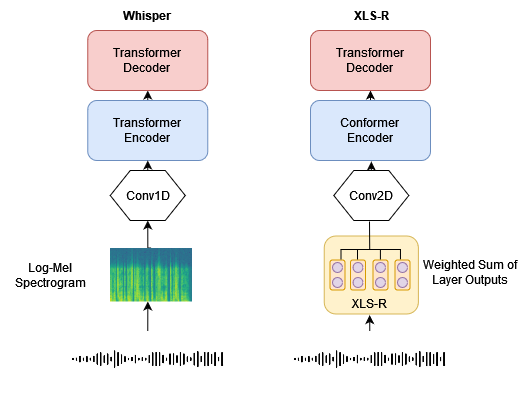}
    \caption{Overview of the two ASR architectures, Whisper (left) and XLS-R (right).}
    \label{fig:asr}
\end{figure}

In order to test ICE, we experiment with two ASR models that are widely used as novel models at this point in time: \textit{Whisper} \cite{radford2023robust} and \textit{XLS-R} \cite{radford2023robust} as seen in Figure \ref{fig:asr}. Both models have been found to perform well on recent speech tasks such as the International Workshop on Speech Translation 2023 \cite{agarwal2023findings}. To be more specific, we use an augmented version of XLS-R that was fine-tuned on multiple languages \cite{chen2023improving} with the hope that it may capture dialect differences. In order to fine-tune the model we use 22 hours of randomly selected audio files (.wav) from the ICE corpus. As a form of validation/development, we used 7.5 hours of files and tested on 9 hours. For Whisper, since it is generally used for what is known as \textit{zero-shot} recognition and does not require fine-tuning, we use the latest version (commit 1838) from OpenAI\footnote{\url{https://github.com/openai/whisper}}.

\section{Results}
\label{sec:results}

Our results are divided into the two tasks presented: Toxicity and ASR. We present our findings with the corresponding metrics. For toxicity, measurements are done using a percentage from 1 to 100\% where 100\% represents full toxicity. For ASR, we use the standard metric known as word-error rate (WER) which measures the number of words correctly predicted by the model. 

\begin{table*}[!htbp]
    \newcolumntype{C}{>{\centering\arraybackslash}X}
    \centering
    \caption{Automatic Speech Recognition results by using word-error rate ($\downarrow$)}
    \label{tab:asr-result}
    \begin{tabularx}{\textwidth}{lCCCC}
        \hline
         & \textbf{Deepcut} &  \textbf{ICE} \\
        \hline
        Whisper Small & 123.5 & 93.8 \\
        XLS-R  & 230.8 & 39.9 \\
        \hline
    \end{tabularx}
\end{table*}

\subsection{Toxicity}
\label{sec:results:toxicity}

In Figure \ref{tab:toxicity-table}, we provide the results of the ETOX toxic evaluation tool on both movies along with the ICE corpus. Toxicity is on par for both languages as expected. While both movies are related to family topics, we do not have a parental rating for Deep Cut. Acrimony is rated `R' by the US administration; therefore, it can be expected to have somewhat more toxic language. Additionally, other factors like diversity in the USA\footnote{\url{https://www.census.gov/newsroom/blogs/random-samplings/2023/05/racial-ethnic-diversity-adults-children.html}} that mark differences between the two countries \textit{could be} considered as important factors. However, it is out of the scope of this paper and saved for future work. 

Measurements for the ICE corpus across all sets: \textit{train}, \textit{development/validation}, and \textit{test} were below 1\% and statistically insignificant. As part of our next iteration, we would like to consider more corpora of different genres and dialects.

\subsection{ASR}
\label{sec:results:asr}

ASR for Nigerian English was remarkably insufficient using the latest techniques. In some cases, the amount of text produced by what can be considered novel techniques introduced words that had no match, causing WERs higher than 100\%. Results for the Whisper and XLS-R approaches are found in Table \ref{tab:asr-result}.  

Our experiments show that for the Nigerian Deep Cut movie, WER in excess of 100\% (124 and 231 for Whisper and XLS-R respectively). Our analysis shows that the multilingual nature of Whisper seems to produce words for the Nigerian English speech into another language spoken in Africa such as Arabic or even Devanagari, a language common in Northern India. At this point in time, we do not have an explanation of why Whisper produces this type of text, one thought is that the dialects from Arabic and Devanagari may be somewhat similar to the Nigerian dialect -- at this point we are clearly assuming and leave verification of intuition for future work.

Whisper fails to recognize the Nigerian speech, with a WER of over 90\%. We found that this is usually because Whisper is unable to properly identify the language being spoken, often incorrectly transcribing the speech into Arabic or Devanagari text. On the other hand, the XLS-R model, despite it being fine-tuned on Nigerian speech, does not perform well either. We believe that this could be due to the lack of Nigerian English in the XLS-R training data. For the ICE corpus, on the other hand, Whisper performs under 100\% with nearly 94\% error which is considered to be remarkably erroneous when compared with other English ASR systems like those created for USA English. XLS-R contrastingly scores quite well compared to all of the other ASR systems with about 40\% WER on the ICE corpus. We consider these findings important and feel that further hyper-parameter search using XLS-R is warranted.

\section{Conclusion}
\label{sec:conclusion}
We conclude our experiments and findings with a clear explanation: \textit{Nollywood movies are great to watch but hard to process}. The experiments performed show that, despite the great advancements in English, the high-resource language used more often for experiments in Artificial Intelligence, low-resource language influence on languages like Nigerian make it more complex to process and build tools for such nations. It is comforting to know that the Nollywood movie along with the formal ICE corpus seem to be less biased and contain less toxicity than their USA counterparts. 

The goal of this paper was to show that Nollywood movies from Nigeria should be considered high-quality movies to watch. Although, if one would like to watch them in their dialect (allbeit English), it may be a while as research has not been advanced much in this area. Additionally, while several African languages like Tamasheq and others are becoming more prevalent in large tasks such as IWSLT 2023\cite{agarwal2023findings}, dialectal tasks should include other dialects such as the English dialect from Nigeria.

\section{Future Work}
\label{sec:future}
We have noted through this article several investigative opportunities which we feel need to be addressed. For example, this work focuses on two mainstream movies and corpora. The next step would be to perform a large-scale search and inclusion of Nigerian corpora. Toxicity and bias can be measured on those corpora and should be compared to more corpora from the USA or other countries. Our systems generated words in Arabic and Devanagari. The next investigations should be to better understand why these systems produce those words for English spoken with a Nigerian dialect.

\begin{figure*}[!htbp]
    \centering
    \includegraphics[width=0.78\textwidth]{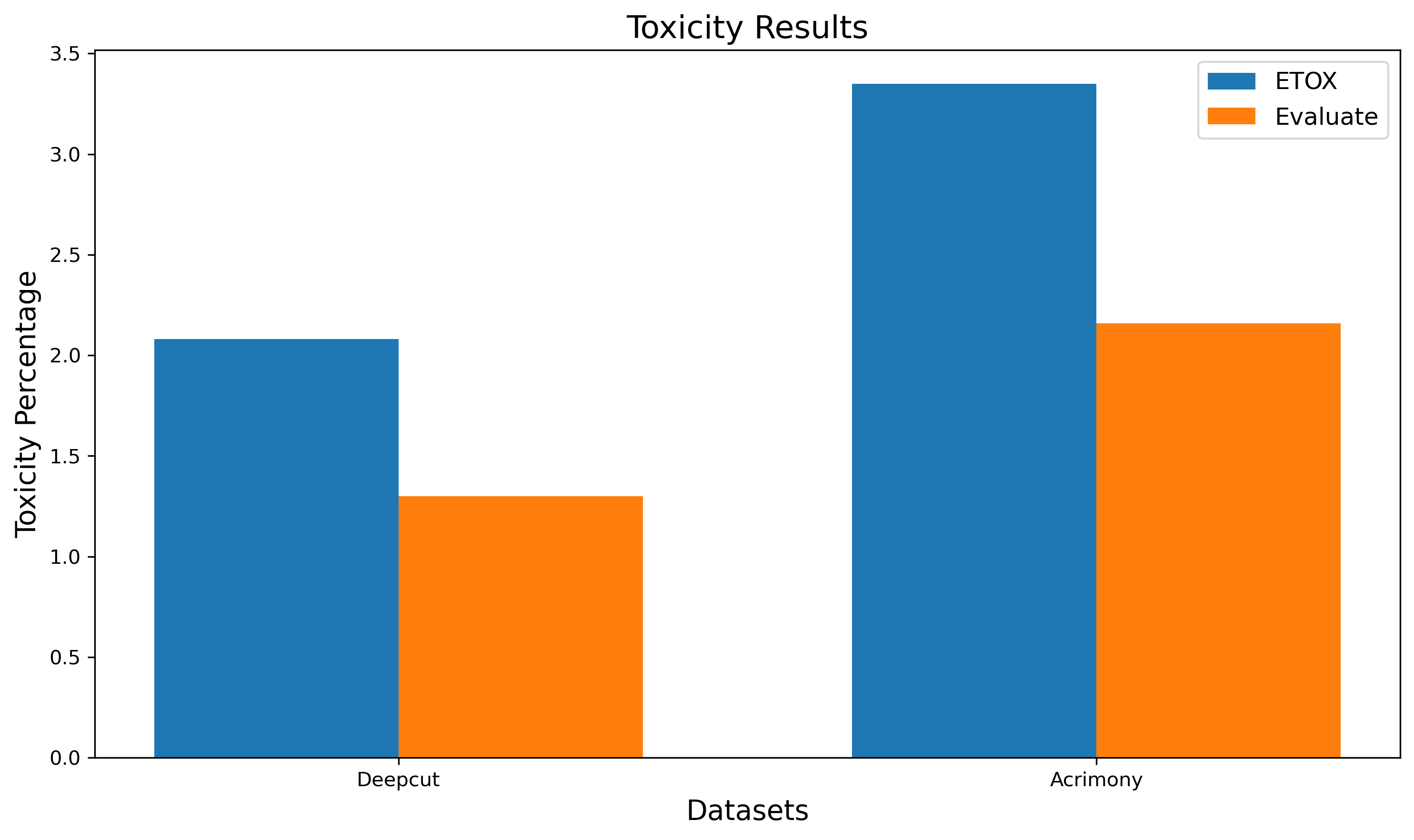}
    \caption{Toxicity results for Deepcut and Acrimony datasets.}
    \label{fig:toxicity-bar-chart}
\end{figure*}

\newpage

\bibliography{acl_latex}
\end{document}